\documentclass[11pt, letterpaper]{article}

\usepackage{amsmath}
\usepackage{amssymb}
\usepackage{amsthm}
\usepackage{graphicx}
\usepackage{hyperref}
\usepackage{url}
\usepackage{algorithm}
\usepackage{algorithmic}
\usepackage{caption}
\usepackage{subcaption}
\usepackage{booktabs}
\usepackage{authblk}
\usepackage{appendix}
\usepackage[utf8]{inputenc}
\usepackage[T1]{fontenc}
\usepackage[english]{babel}
\usepackage{geometry}
\usepackage{xcolor}
\usepackage{fancyhdr}
\usepackage{microtype}
\usepackage{parskip}
\usepackage{tabularx}
\usepackage{enumitem}

\usepackage{titlesec}
\usepackage{indentfirst}

\titleformat{\subsubsection}[hang]{\bfseries\normalsize}{\hspace{0em}}{0pt}{}

\usepackage{float}
\usepackage[numbers,sort]{natbib}
\hypersetup{
    colorlinks=true,
    linkcolor=blue,
    filecolor=magenta,      
    urlcolor=cyan,
    citecolor=red,
}

\newcolumntype{C}{>{\centering\arraybackslash}X} 
\usepackage[normalem]{ulem}

\hypersetup{
    colorlinks=true,
    linkcolor=blue,
    filecolor=magenta,      
    urlcolor=cyan,
    citecolor=red,
}

\definecolor{headercolor}{RGB}{0, 50, 100}
\newcommand*\samethanks[1][\value{footnote}]{\footnotemark[#1]}

\title{LLM-POTUS Score: A Framework of Analyzing Presidential Debates with Large Language Models}

\author[1]{Zhengliang Liu\thanks{Contributed equally}}
\author[1]{Yiwei Li\samethanks}
\author[1]{Oleksandra Zolotarevych}
\author[2]{Rongwei Yang}
\author[1]{Tianming Liu\thanks{Corresponding author}}

\affil[1]{School of Computing, University of Georgia, Athens, GA, USA}
\affil[2]{Department of Mathematics and Statistics, State University of New York at Albany, Albany, NY, USA}

\date{}

\begin{document}

\maketitle

\begin{abstract}
Large language models have demonstrated remarkable capabilities in natural language processing, yet their application to political discourse analysis remains underexplored. This paper introduces a novel approach to evaluating presidential debate performances using LLMs, addressing the longstanding challenge of objectively assessing debate outcomes. We propose a framework that analyzes candidates' "Policies, Persona, and Perspective" (3P) and how they resonate with the "Interests, Ideologies, and Identity" (3I) of four key audience groups: voters, businesses, donors, and politicians. Our method employs large language models to generate the LLM-POTUS Score, a quantitative measure of debate performance based on the alignment between 3P and 3I. We apply this framework to analyze transcripts from recent U.S. presidential debates, demonstrating its ability to provide nuanced, multi-dimensional assessments of candidate performances. Our results reveal insights into the effectiveness of different debating strategies and their impact on various audience segments. This study not only offers a new tool for political analysis but also explores the potential and limitations of using LLMs as impartial judges in complex social contexts. \textbf{In addition, this framework provides individual citizens with an independent tool to evaluate presidential debate performances, which enhances democratic engagement and reduces reliance on potentially biased media interpretations and institutional influence, thereby strengthening the foundation of informed civic participation.}
\end{abstract}

\tableofcontents

\section{Introduction}
The presidential debate stands as a cornerstone of democratic discourse. It offers a rare platform where candidates directly confront each other's ideas, policies, and character~\cite{benoit2013political}. These high-stakes encounters can significantly influence public opinion and, ultimately, election outcomes~\cite{yawn1998presidential,benoit2013political,benoit2003topic}. However, determining the "winner" of a debate is a subjective and contentious process, often clouded by personal biases, media narratives, and the complex interplay of factors that shape voter perceptions.

Traditionally, debate analysis and election forecasting have relied on a combination of public opinion polls, expert commentary, and media interpretations~\cite{brody1991assessing,erikson2019american}. While these approaches provide valuable insights, it is challenging to separate objective analysis from potential biases~\cite{robb2020methodological,d2000media}, limited scope, and the inability to capture the nuanced dynamics of debate performances, especially given the natural challenge posed by the lengthy duration of the debates. The advent of social media and real-time sentiment analysis~\cite{balahur2013sentiment,yoo2018social} has added new dimensions to this field, yet a comprehensive, objective framework for evaluating debate effectiveness remains elusive.

Research consistently demonstrates that human evaluations are inherently susceptible to bias, making it challenging to achieve objectivity and impartiality. Individuals with different political inclinations inevitably interpret the same political debate through a biased lens, with their pre-existing beliefs influencing how they perceive and assess candidates and policies. This is particularly evident in media consumption, where Republicans and Democrats often seek out information that aligns with their views, leading to polarized opinions even when engaging with the same facts. Studies have shown that both sides tend to overestimate or misconstrue issues such as immigration and income inequality, further complicating the pursuit of unbiased assessments~\cite{ecker2024misinformation}.

Moreover, the influence of misinformation, especially during elections, exacerbates these biases. Voters not only misjudge the extent to which misinformation affects others but also allow it to erode their trust in the electoral process. This effect is particularly prominent among Democrats and Independents, whose perception of misinformation's impact on others diminishes their overall satisfaction with democracy~\cite{belcastro2022analyzing}.

In contrast to these human limitations, Large Language Models (LLMs) offer an alternative that is highly professional, unbiased, and efficient in evaluating political content. LLMs, unburdened by emotional or partisan attachments, can provide assessments that are both neutral and informed by a broad set of data. As such, they present a valuable tool for generating fair, efficient, and expert-level evaluations, offering a more balanced perspective compared to human judgment, which is often swayed by personal biases and misinformation~\cite{fiedler1996explaining}.

Recent advancements in artificial intelligence, particularly in the domain of large language models (LLMs)~\cite{liu2023summary,zhou2023comprehensive,zhao2023brain}, offer a promising new approach to this longstanding challenge. LLMs, trained on vast corpora of text data, have demonstrated remarkable capabilities in understanding and generating human-like text across various domains~\cite{liu2023summary}. Their ability to process and analyze complex language patterns, contextual nuances, and semantic relationships~\cite{zhou2023comprehensive,liu2023summary} positions them as potential tools for more systematic and comprehensive debate analysis.

This paper proposes a novel framework for analyzing presidential debates using LLMs, aiming to bridge the gap between subjective human interpretations and data-driven analysis. Our approach is grounded in the understanding that effective debate performance is not merely about policy positions or rhetorical skill, but about how well a candidate's overall presentation resonates with diverse audience segments.

We introduce the concept of "3P-3I alignment" as the cornerstone of our analysis. The 3P represents the key elements of a candidate's debate performance: Policies (the specific proposals and stances), Persona (the consistent image and character projected), and Perspective (the viewpoint or lens through which issues are addressed). The 3I represents the primary concerns of four critical audience groups (voters, businesses, donors, and politicians): Interests (practical benefits or concerns), Ideologies (philosophical or political beliefs), and Identity (personal or group affiliations).

By leveraging the language understanding capabilities of LLMs, we develop the quantitative LLM-POTUS Score that measures how effectively each candidate's 3P aligns with the audience's 3I. The term "LLM-POTUS" is a portmanteau combining "LLM" (large language models) and "POTUS" (President of the United States). This index provides a multi-dimensional score that captures the nuanced effectiveness of debate performances beyond simplistic "win-lose" dichotomies.

Our study not only offers a new tool for political analysis. By applying this framework to recent U.S. presidential debates, we aim to demonstrate its utility in providing insights into debating strategies, audience resonance, and the dynamics of political communication.

More importantly, by offering a tool that individual citizens can use to independently analyze debate performances, the LLM-POTUS Score has the potential to democratize political discourse analysis. This methodology potentially empowers voters to form their own assessments free from the influence of media bias or external commentators.

The following sections will detail our methodology, present our findings from analyzing recent presidential debates, discuss the implications of our results, and explore the broader significance of this approach for political discourse analysis and the evolving role of AI in democratic processes.

\section{Related Work}
There is rising interest in applying large language models to various tasks in political science~\cite{wu2023large,linegar2023large,rodman2024political}. This section provides an overview of relevant studies that have explored the potential of LLMs in political analysis and research.

One of the primary applications of LLMs in political science is the automation of text annotation and classification tasks. LLMs have shown promising results in replacing manual coding efforts, particularly in processing political content. For example, Heseltine and Clemm von Hohenberg~\cite{heseltine2024large} demonstrated that GPT-4~\cite{achiam2023gpt} can accurately code political texts across multiple variables and countries, performing comparably to human experts, especially for shorter texts like tweets. Similarly, Ornstein et al.~\cite{ornstein2022train} found that LLMs can outperform existing automated text classification methods in tasks such as sentiment analysis of political tweets and political ad tone classification, producing results comparable to human crowd-coders at a fraction of the cost.

In addition to classification tasks, LLMs have shown potential in more complex analytical tasks, such as estimating ideological positions. Wu et al.~\cite{wu2023large} proposed a novel framework using LLMs to estimate the latent positions of politicians on various political dimensions. By prompting LLMs to perform pairwise comparisons of lawmakers, they were able to generate scales measuring positions on liberal-conservative ideology, gun control, and abortion rights. O'Hagan and Schein~\cite{o2023measurement} further explored the use of LLMs for ideological scaling of text, demonstrating that these models can provide nuanced interpretations of subtle ideological cues in language.

LLMs also offer new possibilities for generating research content and simulating political scenarios. Linegar et al.~\cite{linegar2023large} discuss how LLMs can be used to generate content for research purposes, such as creating politically biased text for studying social perceptions or simulating multiple humans to replicate human subject studies. This capability opens up new avenues for experimental research in political communication and behavior.

Furthermore, LLMs have shown promise in enhancing our understanding of political speech and discourse. Researchers have used LLMs to analyze the effects of political speech on polarization and to map political speech from less understood domains (like cable news) to established ideological spaces~\cite{linegar2023large}. This application of LLMs provides a systematic way to understand how specific variations in speech affect its persuasiveness and impact.

While these studies have demonstrated the diverse applications of LLMs in political science, our work is the first to apply LLMs directly as judges to evaluate presidential debate performance. By leveraging the language understanding capabilities of LLMs to assess the alignment between candidates' presentations and audience concerns, we introduce a novel approach to quantifying debate effectiveness and extending the use of LLMs in political analysis to a new domain.

\section{Methodology}
\subsection{Framework}
This study introduces a novel approach to evaluating presidential debate performances, the LLM-POTUS Score, which leverages the capabilities and knowledge of large language models.

The LLM-POTUS Score is based on three key dimensions:
\begin{enumerate}
\item \textbf{Policies-Interests Alignment:} How well a candidate's proposed policies resonate with the interests of the electorate.
\item \textbf{Persona-Identity Alignment:} How effectively a candidate's personal brand and image align with the identity of their target voters.
\item \textbf{Perspective-Ideologies Alignment:} How closely a candidate's worldview and ideological stance match those of their supporters.
\end{enumerate}
For each dimension, the LLM assigns a Likert score~\cite{likert1932technique} ranging from 1 to 5:
\begin{itemize}
\item 1: Poor alignment
\item 2: Fair alignment
\item 3: Moderate alignment
\item 4: Good alignment
\item 5: Strong alignment
\end{itemize}
The final LLM-POTUS Score is calculated as the average of these three scores:
\begin{equation}
\text{LLM-POTUS Score} = \frac{\text{Policies-Interests} + \text{Persona-Identity} + \text{Perspective-Ideologies}}{3}
\end{equation}
\begin{table}[h]
\centering
\begin{tabular}{|p{4cm}|p{5cm}|p{5cm}|}
\hline
& \textbf{Trump Supporters} & \textbf{Clinton Supporters} \\
\hline
\textbf{Interests} & Job security, lower taxes, national security & Healthcare access, income equality, social justice \\
\hline
\textbf{Identities} & Working-class whites, conservatives, evangelical Christians & Minorities, women, college-educated professionals \\
\hline
\textbf{Ideologies} & Nationalism, cultural conservatism, anti-globalism & Progressivism, multiculturalism, international cooperation \\
\hline
\end{tabular}
\caption{An Example from 2016: Comparison of Voter Base Characteristics}
\end{table}
To generate these scores, we input the complete debate transcripts into widely-used and highly recognized LLMs: GPT-4~\cite{achiam2023gpt} and Claude 3.5 Sonnet~\cite{anthropicIntroducingClaude} and Llama3.1 405b~\cite{dubey2024llama}. Each model analyzes the content, considering factors such as policy proposals, rhetorical strategies, persona projection, and ideological positioning, relying on their inherent knowledge of U.S. politics and current events.

\subsection{Data}
Our study analyzes U.S. presidential debates from seven election cycles:
\begin{enumerate}
\item \textbf{2000 Presidential Debate:} featuring Al Gore (Democratic) and George W. Bush (Republican)
\begin{itemize}
\item Al Gore: Incumbent Vice President and former U.S. Senator from Tennessee
\item George W. Bush: Governor of Texas and son of former President George H.W. Bush
\end{itemize}
\item \textbf{2004 Presidential Debate:} featuring John Kerry (Democratic) and George W. Bush (Republican)
\begin{itemize}
    \item John Kerry: U.S. Senator from Massachusetts and decorated Vietnam War veteran
    \item George W. Bush: Incumbent President seeking his second term
\end{itemize}

\item \textbf{2008 Presidential Debate:} featuring Barack Obama (Democratic) and John McCain (Republican)
\begin{itemize}
    \item Barack Obama: U.S. Senator from Illinois and former state senator
    \item John McCain: U.S. Senator from Arizona and former Vietnam War prisoner of war
\end{itemize}

\item \textbf{2012 Presidential Debate:} featuring Barack Obama (Democratic) and Mitt Romney (Republican)
\begin{itemize}
    \item Barack Obama: Incumbent President seeking his second term
    \item Mitt Romney: Former Governor of Massachusetts and successful businessman
\end{itemize}

\item \textbf{2016 Presidential Debate:} featuring Hillary Clinton (Democratic) and Donald Trump (Republican)
\begin{itemize}
    \item Hillary Clinton: Former U.S. Secretary of State and former U.S. Senator from New York
    \item Donald Trump: Real estate businessman and television personality with no prior political experience
\end{itemize}

\item \textbf{2020 Presidential Debate:} featuring Joe Biden (Democratic) and Donald Trump (Republican)
\begin{itemize}
    \item Joe Biden: Former Vice President and long-time U.S. Senator from Delaware
    \item Donald Trump: Incumbent President seeking his second term
\end{itemize}

\item \textbf{2024 Presidential Debate:} featuring Kamala Harris (Democratic) and Donald Trump (Republican)
\begin{itemize}
    \item Kamala Harris: Incumbent Vice President and former U.S. Senator from California
    \item Donald Trump: Former President seeking to return to office
\end{itemize}
\end{enumerate}
For each election cycle, we randomly selected one debate for analysis. We obtained the complete, unprocessed debate transcripts for the years 2000 through 2020 directly from the Commission on Presidential Debates\footnote{\url{https://www.debates.org/}}. The transcript for the 2024 debate was sourced from ABC News\footnote{\url{https://abcnews.go.com/amp/Politics/harris-trump-presidential-debate-transcript/story?id=113560542}}. These transcripts include all content from the debates, including moderator comments and questions. No preprocessing or filtering was performed on the transcripts before input into the LLMs.

\subsection{Important Note: Scope of Presented Results in the Main Paper}
While we analyzed all seven election cycles spanning the past 24 years with 3 LLMs, to maintain conciseness in this paper and prevent difficulties in the paper submission process, we present only the results from 2012 to 2024 for GPT-4 and Claude 3.5 Sonnet. The complete set of results, including those from earlier debates and the Llama3.1 405b model, will be compiled separately as an appendix to this study.

\subsection{Important Note: Attribution of Analyses}
It is important to note that all analyses, interpretations, and evaluations of the debate content presented in this study are generated entirely by the LLMs themselves, reflecting their own understanding and perspectives based on the provided transcripts, not those of the authors.

\begin{figure}[t]
\label{pipeline}
\begin{center}
\includegraphics[width=0.9\textwidth]{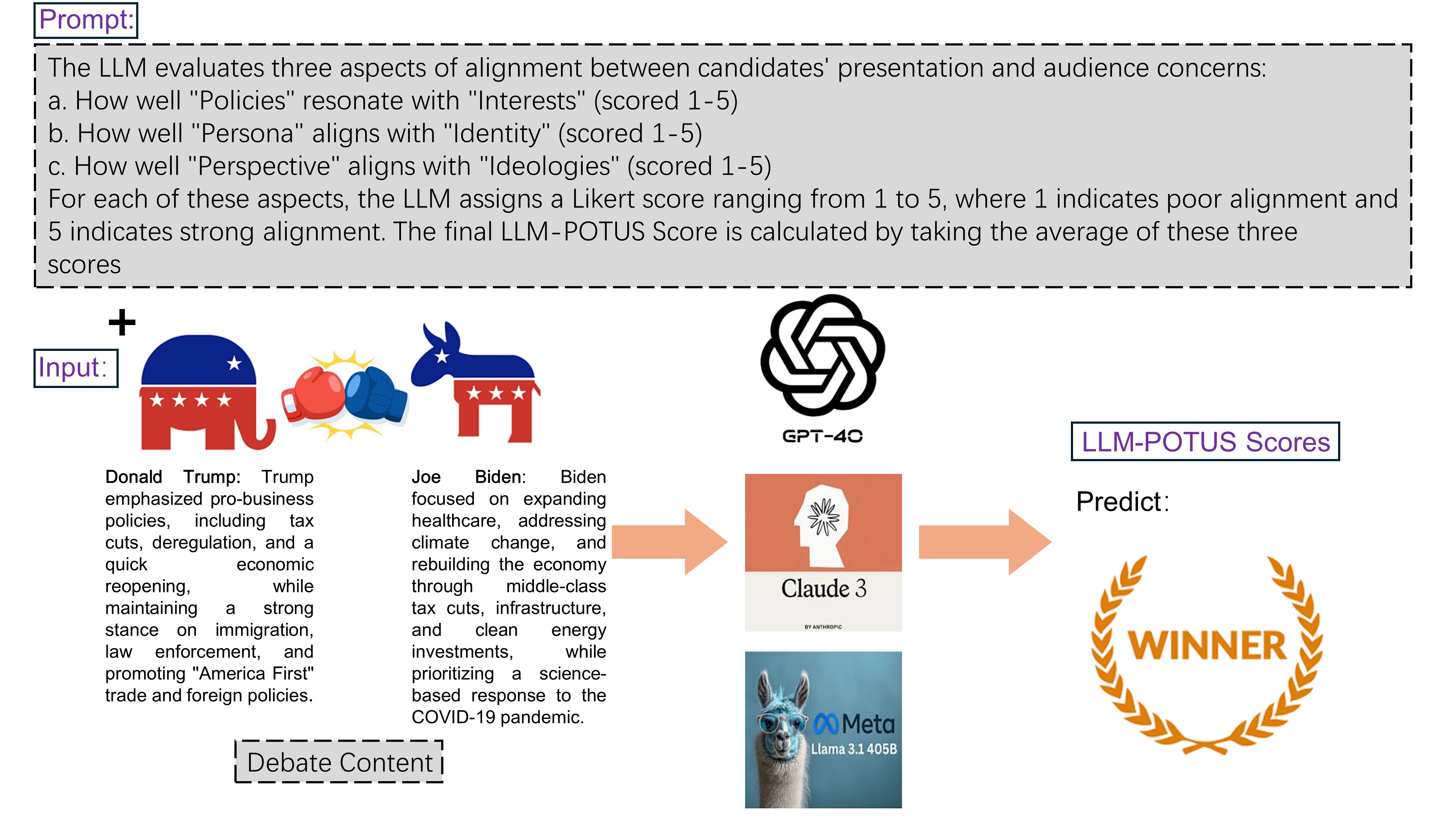}
\end{center}
\caption{The pipeline of analyzing presidential debates with LLMs.}
\end{figure}
\section{Evaluation \& Results}
\subsection{2024 Debate GPT4o}
\subsubsection{Kamala Harris}
\paragraph{Policies-Interests Score: 4/5}
\begin{itemize}
\item Expanding Healthcare: Defended and promised to expand the Affordable Care Act, seeking to ensure universal healthcare access and lower costs for prescription drugs.
\item Progressive Tax Policy: Advocated for higher taxes on the wealthy to fund programs like affordable housing, education, and healthcare expansion.
\item Comprehensive Immigration Reform: Promised to create a path to citizenship for undocumented immigrants while maintaining border security reforms.
\item Climate Change Action: Committed to aggressive climate change policies, including rejoining international agreements like the Paris Accord and investing heavily in renewable energy.
\item Support for Middle-Class Families: Proposed targeted tax cuts for the middle class and increased support for small businesses, including incentives for startups.
\end{itemize}
\paragraph{Persona-Identity Score: 4/5}
\begin{itemize}
\item Experienced Politician: Emphasized her experience as a U.S. Senator and Vice President, framing herself as a capable leader who understands how to govern.
\item Pragmatic Problem-Solver: Presented herself as someone focused on practical solutions to complex issues like healthcare, climate change, and the economy, highlighting her record of legislative work.
\item Compassionate and Inclusive: Positioned herself as an advocate for marginalized communities, particularly in areas of racial justice, healthcare, and economic equity.
\item Strong Communicator: Throughout the debate, she projected a calm, assertive demeanor, often addressing complex issues in a detailed, clear manner.
\item Trailblazer and Advocate for Women: As the first woman of color to run on a major party's presidential ticket, Harris framed herself as a voice for women and minority rights.
\end{itemize}
\paragraph{Perspective-Ideologies Score: 4/5}
\begin{itemize}
\item Progressive: Supported progressive policies in healthcare, climate action, and economic reforms that centered on reducing inequality and expanding social programs.
\item Internationalist: Advocated for a collaborative, multilateral approach to foreign policy, emphasizing the importance of international agreements on trade, climate, and security.
\item Pro-Free Trade (with safeguards): Supported free trade agreements, but with protections for labor and environmental standards, positioning her as a defender of globalization with reforms to benefit American workers.
\item Advocate for Minority Rights: Strongly supported civil rights, focusing on racial justice, voting rights protections, and reforms to the criminal justice system.
\item Believer in Active Government: Consistently argued for a more involved federal government role in providing healthcare, education, and environmental protections.
\end{itemize}
\textbf{Harris's LLM-POTUS Score:} $\frac{4 + 4 + 4}{3} = 4.00$
\subsubsection{Donald Trump}
\paragraph{Policies-Interests Score: 3/5}
\begin{itemize}
\item Repealing and replacing Obamacare: Remained committed to repealing the Affordable Care Act and replacing it with a more market-based solution, though details were sparse.
\item Lowering Taxes: Continued advocating for corporate tax cuts and lower taxes across the board to stimulate business investment and economic growth.
\item Tough Immigration Stance: Promised to use the military and local police to deport millions of undocumented immigrants, expanding on his earlier policies.
\item America First Foreign Policy: Focused on reducing U.S. involvement in international commitments and prioritizing American interests in trade and security.
\item Trade Protectionism: Critical of international trade agreements like NAFTA, advocating for renegotiated deals that favored American workers and industries.
\end{itemize}
\paragraph{Persona-Identity Score: 5/5}
\begin{itemize}
\item Outsider/Non-politician: Continued to position himself as a Washington outsider, railing against career politicians and the establishment.
\item Successful Businessman: Emphasized his background as a businessman capable of running the government more efficiently.
\item Tough/Strong Leader: Projected an image of strength, particularly in law enforcement, national security, and foreign policy, framing himself as the leader America needed in times of uncertainty.
\item "Tells it Like it Is": Maintained his reputation for bluntness, speaking in a direct, often confrontational style.
\item Anti-establishment: Continued his appeal to voters disillusioned with the political elite, positioning himself as the candidate who could challenge entrenched interests.
\end{itemize}
\paragraph{Perspective-Ideologies Score: 4/5}
\begin{itemize}
\item Conservative: Policy positions remained firmly conservative, particularly in economic matters like tax cuts, deregulation, and healthcare reform.
\item Nationalist: Promoted an "America First" vision, emphasizing sovereignty, national pride, and skepticism of global institutions.
\item Protectionist on Trade: Advocated for renegotiating trade deals to protect American workers and industries, opposing multilateral agreements that he argued disadvantaged the U.S.
\item Skeptical of International Commitments: Expressed doubt about international alliances and agreements, preferring a unilateral approach to foreign policy.
\item Law and Order Advocate: Continued his strong stance on law enforcement, supporting police and promising to crack down on protests and civil unrest.
\end{itemize}
\textbf{Trump's LLM-POTUS Score:} $\frac{3 + 5 + 4}{3} = 4.00$

\begin{table}[h]
\centering
\begin{tabular}{|l|c|c|}
\hline
\textbf{Aspect} & \textbf{Donald Trump} & \textbf{Kamala Harris} \\
\hline
Policies - Interests & 3/5 & 4/5 \\
Persona - Identity & 5/5 & 4/5 \\
Perspective - Ideologies & 4/5 & 4/5 \\
\hline
\textbf{Average Score} & 4.00 & 4.00 \\
\hline
\end{tabular}
\caption{2024 LLM-POTUS Scores GPT4o Comparison}
\end{table}

\subsection{2024 Debate Claude3}
\subsubsection{Kamala Harris}
\paragraph{Policies-Interests Score: 4/5}
\begin{itemize}
\item \textbf{Economic plan:} Harris outlined specific policies for job creation, small business support, and affordable housing, addressing key voter concerns.
\item \textbf{Healthcare:} She defended and proposed expanding the Affordable Care Act, addressing healthcare access concerns.
\item \textbf{Climate change:} Harris acknowledged climate change as a serious issue and discussed investments in clean energy.
\item \textbf{Social issues:} She addressed racial justice and women's rights, aligning with progressive voters' interests.
\item \textbf{Foreign policy:} Harris emphasized support for Ukraine and NATO, appealing to voters favoring traditional alliances.
\end{itemize}
\paragraph{Persona-Identity Score: 4/5}
\begin{itemize}
\item \textbf{Experience:} Harris emphasized her diverse political experience while positioning herself as a fresh alternative.
\item \textbf{Composure:} She generally maintained a calm demeanor, though occasionally showed frustration with Trump's attacks.
\item \textbf{Empathy:} Harris attempted to connect with average Americans by sharing personal stories and focusing on kitchen-table issues.
\item \textbf{Unity message:} She positioned herself as a unifier who could work across party lines.
\item \textbf{New generation:} Harris presented herself as a bridge between established leadership and new ideas.
\end{itemize}
\paragraph{Perspective-Ideologies Score: 4/5}
\begin{itemize}
\item \textbf{Progressive policies:} Harris advocated for expanding social programs and addressing income inequality, aligning with progressive ideologies.
\item \textbf{Moderate appeal:} She distanced herself from some far-left policies, attempting to appeal to moderates.
\item \textbf{International cooperation:} Her stance on supporting Ukraine and NATO aligns with liberal internationalist ideologies.
\item \textbf{Role of government:} Harris's policy proposals generally involved an active role for government in solving problems.
\item \textbf{Social justice:} Her acknowledgment of systemic racism and call for reforms aligned with progressive racial justice ideologies.
\end{itemize}
\textbf{Harris's LLM-POTUS Score:} $\frac{4 + 4 + 4}{3} = 4.00$

\subsubsection{Donald Trump}
\paragraph{Policies-Interests Score: 3/5}
\begin{itemize}
\item \textbf{Economy focus:} Trump emphasized job creation, renegotiating trade deals, and his past economic record, appealing to voters concerned about the economy.
\item \textbf{Immigration:} He maintained a hardline stance on immigration and border security, aligning with his base's interests.
\item \textbf{Foreign policy:} Trump criticized existing alliances and emphasized his ability to negotiate with world leaders.
\item \textbf{Healthcare:} He criticized Obamacare but struggled to articulate a clear alternative plan.
\item \textbf{Climate change:} Trump downplayed environmental concerns, potentially alienating environmentally conscious voters.
\end{itemize}
\paragraph{Persona-Identity Score: 4/5}
\begin{itemize}
\item \textbf{Outsider image:} Trump consistently portrayed himself as fighting against the political establishment.
\item \textbf{Confrontational style:} His aggressive debating style and frequent interruptions reinforced his image as a "fighter."
\item \textbf{Confidence:} He projected strong self-assurance in his abilities and past accomplishments.
\item \textbf{America First:} Trump consistently emphasized putting American interests first in all policies.
\item \textbf{Nostalgia:} He appealed to voters' desire to return to what he portrayed as a better past.
\end{itemize}
\paragraph{Perspective-Ideologies Score: 4/5}
\begin{itemize}
\item \textbf{Economic nationalism:} Trump's views on trade and jobs aligned with economic nationalist ideologies.
\item \textbf{Law and order:} His stance on immigration and crime reflected conservative law enforcement ideologies.
\item \textbf{Limited government:} He criticized regulations and high taxes, aligning with conservative economic ideology.
\item \textbf{Nationalism:} Trump's foreign policy comments reflected a nationalist, isolationist ideology.
\item \textbf{Anti-establishment:} He consistently positioned himself against political elites and the status quo.
\end{itemize}
\begin{table}[h]
\centering
\begin{tabular}{|l|c|c|}
\hline
\textbf{Aspect} & \textbf{Donald Trump} & \textbf{Kamala Harris} \\
\hline
Policies - Interests & 3/5 & 4/5 \\
Persona - Identity & 4/5 & 4/5 \\
Perspective - Ideologies & 4/5 & 4/5 \\
\hline
\textbf{Average Score} & 3.67 & 4.00 \\
\hline
\end{tabular}
\caption{2024 LLM-POTUS Claude3 Scores Comparison}
\end{table}

\subsection{2020 Debate GPT4o}
\subsubsection{Joe Biden}
\paragraph{Policies-Interests Score: 4.5/5}
\begin{itemize}
\item COVID-19 Response: Emphasized stricter measures to combat COVID-19, including a nationwide mask mandate and coordinated efforts to provide protective equipment and support for testing.
\item Healthcare: Advocated for expanding the Affordable Care Act (Obamacare) and introducing a public option for health insurance.
\item Economic Recovery: Focused on middle- and working-class Americans, proposing investments in infrastructure, clean energy, and education.
\item Climate Change: Called for aggressive action on climate change, including a goal to reach net-zero emissions by 2050.
\item Tax Policy: Planned to reverse Trump's tax cuts for the wealthy and increase taxes on individuals making over \$400,000 annually.
\item Racial Inequality and Police Reform: Acknowledged systemic racism in the U.S. and supported police reforms to increase accountability.
\end{itemize}
\paragraph{Persona-Identity Score: 4/5}
\begin{itemize}
\item Experienced and Steady Leader: Emphasized his extensive experience in public service.
\item Empathetic and Compassionate: Projected a persona of empathy, often referencing his own personal losses.
\item Uniter and Restorer of Civility: Central message was one of unity and restoring decency to the presidency.
\item Calm and Focused: Tried to maintain a calm and collected demeanor throughout the debate.
\item Pragmatic Problem-Solver: Presented himself as a pragmatic leader focused on addressing immediate problems.
\end{itemize}
\paragraph{Perspective-Ideologies Score: 4/5}
\begin{itemize}
\item Progressive Economic and Healthcare Policies: Supported expanding the Affordable Care Act and rebuilding the economy from the bottom up.
\item COVID-19 Response Based on Science: Called for a more robust, science-driven response to the pandemic.
\item International Cooperation and Multilateralism: Advocated for a return to international alliances and diplomacy.
\item Environmental Protection and Climate Action: Supported re-entering the Paris Climate Agreement and proposed aggressive investments in clean energy.
\item Racial Justice and Police Reform: Acknowledged systemic racism and supported police reform to increase accountability.
\end{itemize}
\textbf{Biden's LLM-POTUS Score:} $\frac{4.5 + 4 + 4}{3} = 4.17$
\subsubsection{Donald Trump}
\paragraph{Policies-Interests Score: 3.5/5}
\begin{itemize}
\item COVID-19 Response: Defended his administration's handling of the pandemic, emphasizing the rapid development of vaccines and the reopening of the economy.
\item Economic Recovery: Highlighted the economic recovery under his administration, favoring keeping the economy open and reducing regulations.
\item Healthcare: Argued for the repeal of the Affordable Care Act, claiming it was ineffective and expensive.
\item Tax Policy: Supported maintaining the tax cuts introduced during his presidency, particularly for corporations and high-income earners.
\item Law and Order: Emphasized the need for law and order, opposing movements to defund the police.
\item Climate Policy: Critical of climate change regulations, emphasizing energy independence and the importance of the fossil fuel industry.
\end{itemize}
\paragraph{Persona-Identity Score: 3.5/5}
\begin{itemize}
\item Outsider and Non-Politician: Continued to emphasize his identity as a political outsider.
\item Tough and Combative Leader: Projected an aggressive and confrontational persona.
\item Populist and Champion of the "Forgotten": Positioned himself as the voice of the "forgotten" Americans.
\item Direct and Unfiltered: Maintained a blunt and unfiltered communication style.
\item Law-and-Order Advocate: Framed himself as the candidate who would restore "law and order".
\end{itemize}
\paragraph{Perspective-Ideologies Score: 3.5/5}
\begin{itemize}
\item Pro-Business Economic Approach: Focused on deregulation, tax cuts, and boosting domestic manufacturing.
\item Critic of Obamacare and Healthcare Reform: Continued to call for the repeal of Obamacare.
\item Nationalist and "America First" Foreign Policy: Doubled down on his "America First" foreign policy.
\item COVID-19 Response Focused on Reopening: Defended his administration's handling of the pandemic, focusing on reopening the economy.
\item Anti-Immigration and Strong Border Policies: Maintained his hardline stance on immigration.
\end{itemize}
\textbf{Trump's LLM-POTUS Score:} $\frac{3.5 + 3.5 + 3.5}{3} = 3.50$

\begin{table}[h]
\centering
\begin{tabular}{|l|c|c|}
\hline
\textbf{Aspect} & \textbf{Joe Biden} & \textbf{Donald Trump} \\
\hline
Policies - Interests & 4.5/5 & 3.5/5 \\
Persona - Identity & 4/5 & 3.5/5 \\
Perspective - Ideologies & 4/5 & 3.5/5 \\
\hline
\textbf{Average Score} & 4.17 & 3.50 \\
\hline
\end{tabular}
\caption{2020 LLM-POTUS Scores GPT4o Comparison}
\end{table}
\subsection{2020 Debate Claude3}
\subsubsection{Donald Trump}
\paragraph{Policies-Interests Score: 3/5}
\begin{itemize}
\item Economic focus: Emphasized pre-pandemic economic record, criticized Biden's tax plans.
\item Law and order: Stressed support for law enforcement, criticized "radical left" policies.
\item Healthcare: Claimed to protect pre-existing conditions, lacked specific policy details.
\item Climate change: Downplayed environmental concerns.
\item COVID-19 response: Defended administration's actions, no clear future plan.
\end{itemize}
\paragraph{Persona-Identity Score: 4/5}
\begin{itemize}
\item Combative style: Aggressive debating, frequent interruptions.
\item "Strongman" image: Projected confidence in unilateral problem-solving.
\item Anti-establishment stance: Portrayed as fighting entrenched political interests.
\item Business acumen: Emphasized business background in economic discussions.
\end{itemize}
\paragraph{Perspective-Ideologies Score: 4/5}
\begin{itemize}
\item Conservative judiciary: Highlighted Supreme Court nominations.
\item Limited government: Criticized regulations and high taxes.
\item Law enforcement support: Strong backing of police.
\item Climate skepticism: Skeptical stance on climate change.
\item America First: Nationalist foreign policy stance.
\end{itemize}
\textbf{Trump's LLM-POTUS Score:} $\frac{3 + 4 + 4}{3} = 3.67$
\subsubsection{Joe Biden}
\paragraph{Policies-Interests Score: 4/5}
\begin{itemize}
\item Healthcare: Defended and proposed expanding Affordable Care Act.
\item Climate plan: Outlined clean energy goals, rejoining Paris Agreement.
\item COVID-19 response: Criticized Trump's handling, offered own plan.
\item Economic recovery: Discussed job creation and small business support.
\item Racial justice: Addressed systemic racism, proposed reforms.
\end{itemize}
\paragraph{Persona-Identity Score: 3/5}
\begin{itemize}
\item Experienced leader: Emphasized long political career and Obama administration experience.
\item Empathy: Attempted to connect with average Americans through personal stories.
\item Unity message: Positioned as a unifier across party lines.
\item Composure challenges: Occasional loss of cool, undermining elder statesman image.
\end{itemize}
\paragraph{Perspective-Ideologies Score: 4/5}
\begin{itemize}
\item Progressive policies: Advocated expanding healthcare access, addressing climate change.
\item Moderate appeal: Distanced from some far-left policies.
\item International cooperation: Supported rejoining international agreements.
\item Role of government: Proposed active government role in problem-solving.
\item Racial justice: Acknowledged systemic racism, called for reforms.
\end{itemize}
\textbf{Biden's LLM-POTUS Score:} $\frac{4 + 3 + 4}{3} = 3.67$
Both candidates maintained strong alignment with their core bases while attempting to appeal to broader audiences. The debate's contentious nature often hindered full articulation of positions and consistent persona maintenance, affecting scores in various ways.
\begin{table}[h]
\centering
\begin{tabular}{|l|c|c|}
\hline
\textbf{Aspect} & \textbf{Donald Trump} & \textbf{Joe Biden} \\
\hline
Policies - Interests & 3/5 & 4/5 \\
Persona - Identity & 4/5 & 3/5 \\
Perspective - Ideologies & 4/5 & 4/5 \\
\hline
\textbf{Average Score} & 3.67 & 3.67 \\
\hline
\end{tabular}
\caption{2020 LLM-POTUS Claude3 Scores Comparison}
\end{table}
\subsection{2016 Debate GPT4o}
\subsubsection{Hillary Clinton}
\paragraph{Policies-Interests Score: 4.5/5}
\begin{itemize}
\item Healthcare: Strongly defended the Affordable Care Act (Obamacare) and proposed to build on it.
\item Economic Plan: Proposed an economy that works for everyone, not just those at the top.
\item National Security: Advocated for a strong international coalition to fight terrorism, particularly ISIS.
\item Immigration: Supported comprehensive immigration reform, including a pathway to citizenship for undocumented immigrants.
\item Gun Control: Supported stronger gun control measures, such as universal background checks.
\item Climate Change: Advocated for the U.S. to take a leading role in addressing climate change.
\end{itemize}
\paragraph{Persona-Identity Score: 4/5}
\begin{itemize}
\item Experienced Politician and Public Servant: Positioned herself as a seasoned and capable leader.
\item Calm and Composed: Maintained a calm, measured demeanor throughout the debate.
\item Policy Expert and Pragmatist: Presented herself as someone deeply knowledgeable about policy details.
\item Inclusive and Compassionate Leader: Portrayed herself as a candidate for all Americans.
\item Prepared and Reliable: Emphasized her readiness to lead and handle crises.
\end{itemize}
\paragraph{Perspective-Ideologies Score: 4.5/5}
\begin{itemize}
\item Progressive Domestic Agenda: Focused on expanding access to healthcare, improving education, and raising wages.
\item Defender of Obamacare: Strongly supported the Affordable Care Act and proposed expanding it.
\item Pro-Diplomacy and International Cooperation: Foreign policy perspective rooted in diplomacy and coalition-building.
\item Tough but Measured on National Security: Supported a robust national security strategy while emphasizing diplomacy.
\item Environmental Protection and Climate Action: Advocated for strong action on climate change.
\end{itemize}
\textbf{Clinton's LLM-POTUS Score:} $\frac{4.5 + 4 + 4.5}{3} = 4.33$
\subsubsection{Donald Trump}
\paragraph{Policies-Interests Score: 4/5}
\begin{itemize}
\item Healthcare: Argued for repealing and replacing Obamacare, calling it a failure that raised healthcare costs.
\item Economic Plan: Focused on tax cuts for individuals and corporations.
\item National Security: Emphasized his commitment to defeating ISIS and criticized Clinton's handling of foreign policy.
\item Immigration: Reiterated his plan to build a wall along the southern border and make Mexico pay for it.
\item Trade: Critical of trade deals like NAFTA, which he said hurt American workers.
\item Law and Order: Emphasized the importance of restoring law and order in American cities.
\end{itemize}
\paragraph{Persona-Identity Score: 3.5/5}
\begin{itemize}
\item Outsider and Non-Politician: Positioned himself as a political outsider, not beholden to the traditional political establishment.
\item Tough and Direct: Communication style was blunt and confrontational.
\item Nationalist and Anti-Establishment: Framed his campaign as a revolt against a corrupt political system.
\item Populist and Protector of the "Forgotten": Aimed to appeal to working-class Americans.
\item Aggressive and Unfiltered: Debate persona was aggressive and unfiltered, often interrupting Clinton.
\end{itemize}
\paragraph{Perspective-Ideologies Score: 4/5}
\begin{itemize}
\item Economic Nationalism and Protectionism: Focused on renegotiating trade deals and imposing tariffs.
\item Repeal and Replace Obamacare: Vowed to repeal the Affordable Care Act.
\item "Law and Order" Advocate: Emphasized a strong stance on law and order.
\item Isolationist and America First in Foreign Policy: Advocated for a more isolationist approach to foreign policy.
\item Climate Change Skeptic and Pro-Fossil Fuels: Skeptical of climate change and opposed environmental regulations.
\end{itemize}
\textbf{Trump's LLM-POTUS Score:} $\frac{4 + 3.5 + 4}{3} = 3.83$

\begin{table}[h]
\centering
\begin{tabular}{|l|c|c|}
\hline
\textbf{Aspect} & \textbf{Donald Trump} & \textbf{Hillary Clinton} \\
\hline
Policies - Interests & 4/5 & 4.5/5 \\
Persona - Identity & 3.5/5 & 4/5 \\
Perspective - Ideologies & 4/5 & 4.5/5 \\
\hline
\textbf{Average Score} & 3.83 & 4.33 \\
\hline
\end{tabular}
\caption{2016 LLM-POTUS Scores GPT4o Comparison }
\end{table}
\subsection{2016 Debate Claude3}
\subsubsection{Hillary Clinton}
\paragraph{Policies-Interests Score: 4.5/5}
\begin{itemize}
\item Economy: Proposed investments in infrastructure, manufacturing, clean energy, and small businesses to create jobs and boost economic growth. Clinton stated, "We also have to make the economy fairer. That starts with raising the national minimum wage and also guarantee, finally, equal pay for women's work."
\item Healthcare: Defended and proposed expanding the Affordable Care Act (Obamacare). She emphasized, "I want to make sure we're getting the cost down and costs have come down under the Affordable Care Act."
\item Climate Change: Emphasized the importance of addressing climate change and investing in clean energy. Clinton said, "We can deploy a half a billion more solar panels. We can have enough clean energy to power every home. We can build a new modern electric grid."
\item Tax Policy: Advocated for raising taxes on the wealthy and closing corporate loopholes. She argued, "I think building the middle class, investing in the middle class, making college debt-free so more young people can get their education, helping people refinance their debt from college at a lower rate. Those are the kinds of things that will really boost the economy."
\item Gun Control: Supported stronger gun control measures, including comprehensive background checks. Clinton stated, "We have to make sure that people who are dangerous, people who are violent or have a history of domestic abuse or stalking, do not get guns."
\item Foreign Policy: Emphasized the importance of maintaining international alliances and using diplomacy to address global challenges. She said, "We need to be more precise in how we talk about these issues. People around the world follow our presidential campaigns so closely, trying to get hints about what we will do."
\end{itemize}
\paragraph{Persona-Identity Score: 4/5}
\begin{itemize}
\item Experienced Leader: Frequently highlighted her experience as Secretary of State and Senator. Clinton said, "I think my strongest asset, maybe by far, is my temperament. I have a much better temperament than he has."
\item Policy Expert: Demonstrated detailed knowledge on various issues, often providing specific plans and numbers. This was evident in her detailed responses on economic policy and healthcare.
\item Composed and Prepared: Maintained a calm demeanor throughout the debate, appearing well-prepared for questions and attacks. She often responded to Trump's accusations with "Please, fact checkers, get to work."
\item Champion for Women and Minorities: Positioned herself as an advocate for women's rights and racial equality. She criticized Trump's past comments about women and minorities.
\item Pragmatic Problem-Solver: Presented herself as someone who could work across party lines to achieve results. Clinton emphasized, "I've been a senator, I've been secretary of state. I've been first lady. I've worked across the aisle to get things done."
\end{itemize}
\paragraph{Perspective-Ideologies Score: 4/5}
\begin{itemize}
\item Progressive Domestic Agenda: Supported expanding social programs and addressing income inequality. She stated, "I don't think top-down works in America. I think building the middle class, investing in the middle class, that's what will make our economy grow."
\item Internationalist: Advocated for strong international alliances and multilateral approach to global issues. Clinton emphasized, "We need to work more closely with our allies and that's something that Donald has been very dismissive of."
\item Believer in Active Government: Proposed government solutions to various social and economic challenges, including healthcare reform and infrastructure investment.
\item Environmental Advocate: Strongly supported action on climate change and investment in clean energy. She said, "I think it's real. I think science is real."
\item Social Liberal: Defended women's rights, racial equality, and LGBTQ+ rights. Clinton criticized Trump's past comments and policies regarding these issues.
\end{itemize}
\textbf{Clinton's LLM-POTUS Score:} $\frac{4.5 + 4 + 4}{3} = 4.17$
\subsubsection{Donald Trump}
\paragraph{Policies-Interests Score: 4/5}
\begin{itemize}
\item Economy: Promised to create jobs through tax cuts, deregulation, and renegotiating trade deals. Trump stated, "Under my plan, I'll be reducing taxes tremendously, from 35 percent to 15 percent for companies, small and big businesses."
\item Immigration: Proposed building a wall on the Mexican border and increasing deportations. He emphasized, "We have to stop our jobs from being stolen from us. We have to stop our companies from leaving the United States."
\item Trade: Criticized NAFTA and other trade deals, promising to renegotiate for better terms. Trump said, "NAFTA is the worst trade deal maybe ever signed anywhere, but certainly ever signed in this country."
\item Tax Policy: Proposed significant tax cuts for businesses and individuals. He argued, "I'm getting rid of the carried interest provision. And if you really look, it's not a tax -- it's really not a great thing for the wealthy. It's a great thing for the middle class."
\item Foreign Policy: Advocated for an "America First" approach, criticizing current alliances and commitments. Trump stated, "I think NATO is obsolete. NATO is not taking care of terror."
\item Law and Order: Emphasized the need for tougher law enforcement and support for police. He said, "We need law and order. If we don't have it, we're not going to have a country."
\end{itemize}
\paragraph{Persona-Identity Score: 4/5}
\begin{itemize}
\item Outsider/Non-politician: Consistently portrayed himself as an outsider to the political establishment. Trump emphasized, "It's all words, it's all sound bites. I built an unbelievable company."
\item Successful Businessman: Emphasized his business experience as qualification for presidency. He stated, "I have a tremendous company. I have tremendous income."
\item Strong and Decisive Leader: Projected an image of strength and decisiveness. Trump said, "We need a very strong leader, and we need a very, very strong leadership."
\item "Tells it Like it Is": Used direct, often confrontational language, appealing to voters frustrated with political correctness. This was evident in his frequent interruptions and direct attacks on Clinton.
\item America First Advocate: Positioned himself as a champion for American interests above all else. Trump emphasized, "I'm going to make America great again. We have a country that's doing so badly."
\end{itemize}
\paragraph{Perspective-Ideologies Score: 4/5}
\begin{itemize}
\item Economic Nationalism: Focused on protecting American jobs and industries from foreign competition. Trump stated, "Our jobs are fleeing the country. They're going to Mexico. They're going to many other countries."
\item Anti-Establishment: Criticized both Democratic and Republican establishments. He said, "The problem is, you talk, but you don't get anything done, Hillary. You don't."
\item Law and Order Conservatism: Emphasized tough-on-crime policies and support for law enforcement. Trump argued, "We have gangs roaming the street. And in many cases, they're illegally here, illegal immigrants. And they have guns."
\item Skepticism of Climate Change: Questioned the scientific consensus on climate change. He said, "I did not say that. I do not say that," when Clinton accused him of calling climate change a hoax.
\item Unilateralism in Foreign Policy: Advocated for a more self-interested approach to international relations. Trump emphasized, "We cannot be the policemen of the world. We cannot protect countries all over the world where they're not paying us what we need."
\end{itemize}
\textbf{Trump's LLM-POTUS Score:} $\frac{4 + 4 + 4}{3} = 4.00$

\begin{table}[h]
\centering
\begin{tabular}{|l|c|c|}
\hline
\textbf{Aspect} & \textbf{Donald Trump} & \textbf{Hillary Clinton} \\
\hline
Policies - Interests & 4/5 & 4.5/5 \\
Persona - Identity & 4/5 & 4/5 \\
Perspective - Ideologies & 4/5 & 4/5 \\
\hline
\textbf{Average Score} & 4 & 4.17 \\
\hline
\end{tabular}
\caption{2016 LLM-POTUS Claude3 Scores Comparison}
\end{table}
\subsection{2012 Debate GPT4o}
\subsubsection{Barack Obama}
\paragraph{Policies-Interests Score: 4.5/5}
\begin{itemize}
\item Healthcare: Defended the Affordable Care Act (Obamacare) as a significant achievement.
\item Economic Plan: Advocated for a balanced approach to economic recovery, focusing on job creation through infrastructure projects, clean energy investments, and education.
\item Education: Emphasized the importance of investing in education, particularly through expanding access to community colleges and increasing Pell Grants.
\item Energy Policy: Supported a balanced energy policy, including investments in renewable energy like wind, solar, and biofuels, while also promoting increased production of oil and natural gas.
\item Tax Policy: Proposed raising taxes on the wealthiest Americans, arguing that the rich should pay their fair share to help reduce the deficit.
\item Medicare and Social Security: Defended Medicare and Social Security, rejecting Romney's proposal to turn Medicare into a voucher system.
\end{itemize}
\paragraph{Persona-Identity Score: 4/5}
\begin{itemize}
\item Experienced and Steady Leader: Positioned himself as a leader with four years of experience as president, having managed the country through a severe economic crisis.
\item Calm and Pragmatic: Projected a calm, pragmatic demeanor, focused on presenting himself as a leader who could handle crises thoughtfully and without panic.
\item Inclusive and Hopeful: Framed himself as a president for all Americans, seeking to unite the country across political and social divides.
\item Focused on Middle-Class Advocacy: Positioned himself as a strong advocate for middle-class families, emphasizing economic security and access to affordable healthcare and education.
\item Defender of Key Achievements: Defended major legislative achievements, such as the Affordable Care Act and financial reforms like Dodd-Frank.
\end{itemize}
\paragraph{Perspective-Ideologies Score: 4.5/5}
\begin{itemize}
\item Progressive Economic Policies: Focused on continuing recovery efforts after the financial crisis, supporting middle-class tax cuts and investing in education, clean energy, and infrastructure.
\item Healthcare Reform Advocate: Strongly defended the Affordable Care Act, viewing healthcare as a fundamental right and pushing for expanding access to affordable care.
\item Balanced Approach to Deficit Reduction: Emphasized a balanced approach to reducing the deficit, combining spending cuts with revenue increases.
\item Diplomatic and Multilateral in Foreign Policy: Focused on diplomacy, rebuilding international alliances, and responsibly ending wars.
\item Clean Energy and Environmental Protection: Promoted investment in clean energy and efforts to combat climate change.
\end{itemize}
\textbf{Obama's LLM-POTUS Score:} $\frac{4.5 + 4 + 4.5}{3} = 4.33$
\subsubsection{Mitt Romney}
\paragraph{Policies-Interests Score: 4/5}
\begin{itemize}
\item Healthcare: Advocated for the repeal of Obamacare, arguing that it was a federal overreach and too costly.
\item Economic Plan: Focused on reducing taxes, deregulation, and cutting government spending. Proposed a 20
\item Job Creation: Proposed a five-point plan focused on energy independence, education reform, promoting free trade, reducing the national debt, and supporting small businesses.
\item Energy Policy: Promoted an "all-of-the-above" energy strategy, focused on increasing domestic oil, gas, and coal production.
\item Tax Policy: Advocated for lowering individual tax rates and reducing the corporate tax rate to make the U.S. more competitive globally.
\item Medicare and Social Security: Supported reforming Medicare, including offering future retirees the option of choosing between traditional Medicare or a private plan through a voucher system.
\end{itemize}
\paragraph{Persona-Identity Score: 3.5/5}
\begin{itemize}
\item Successful Businessman and Job Creator: Leaned heavily on his experience as a successful businessman, portraying himself as someone who could use his private-sector background to fix the economy.
\item Decisive and Strong Leader: Projected a decisive and confident persona, positioning himself as someone who could lead with conviction and take swift action.
\item Problem-Solver: Framed himself as a practical problem-solver, someone who could make government work more efficiently and cut through bureaucratic inefficiencies.
\item Clear and Direct Communicator: Aimed to communicate his ideas in a clear, straightforward manner, often critiquing Obama for being too complex or nuanced in his policy explanations.
\item Defender of Conservative Values: Portrayed himself as a defender of traditional conservative values, particularly in relation to economic issues.
\end{itemize}
\paragraph{Perspective-Ideologies Score: 4/5}
\begin{itemize}
\item Pro-Market Economic Approach: Economic platform focused on reducing taxes across the board, cutting regulations, and reducing government spending.
\item Opposition to Obamacare: Critical of the Affordable Care Act and promised to repeal it if elected, advocating for a more market-based solution to healthcare.
\item Aggressive Deficit Reduction: Pushed for more aggressive deficit reduction through significant cuts to government spending.
\item Strong National Defense and Foreign Policy: Advocated for a stronger U.S. military presence globally and a more assertive foreign policy.
\item Traditional Energy Development: Favored an "all of the above" energy strategy but emphasized traditional energy sources like coal, oil, and natural gas.
\end{itemize}
\textbf{Romney's LLM-POTUS Score:} $\frac{4 + 3.5 + 4}{3} = 3.83$

\begin{table}[h]
\centering
\begin{tabular}{|l|c|c|}
\hline
\textbf{Aspect} & \textbf{Barack Obama} & \textbf{Mitt Romney} \\
\hline
Policies - Interests & 4.5/5 & 4/5 \\
Persona - Identity & 4/5 & 3.5/5 \\
Perspective - Ideologies & 4.5/5 & 4/5 \\
\hline
\textbf{Average Score} & 4.33 & 3.83 \\
\hline
\end{tabular}
\caption{2012 LLM-POTUS Scores GPT4o Comparison}
\end{table}

\subsection{2012 Debate Claude3}
\subsubsection{Barack Obama}
\paragraph{Policies-Interests Score: 4/5}
\begin{itemize}
\item Invest in education and training: Proposed hiring 100,000 new math and science teachers.
\item Develop new energy sources: Highlighted investment in renewable energy and increased oil/gas production.
\item Change tax code to help small businesses: Proposed lowering corporate tax rate for manufacturing.
\item Reduce deficit in a balanced way: Advocated for mix of spending cuts and revenue increases.
\item Improve healthcare through Affordable Care Act: Defended ACA benefits.
\item Maintain Social Security and Medicare: Stressed preserving these programs.
\end{itemize}
\paragraph{Persona-Identity Score: 4/5}
\begin{itemize}
\item Experienced leader: Drew on four years as president to demonstrate capabilities.
\item Champion of the middle class: Framed policies in terms of middle-class benefits.
\item Pragmatic problem-solver: Presented as someone who could work across party lines.
\item Defender of his administration's accomplishments: Highlighted first-term positive outcomes.
\end{itemize}
\paragraph{Perspective-Ideologies Score: 5/5}
\begin{itemize}
\item Progressive: Policy proposals aligned with progressive ideals.
\item Believer in active government role: Argued for government involvement in various areas.
\item Internationalist: Suggested commitment to international cooperation.
\item Advocate for balanced approach to deficit reduction: Emphasized mix of cuts and revenue increases.
\end{itemize}
\textbf{Obama's LLM-POTUS Score:} $\frac{4 + 4 + 5}{3} = 4.33$
\subsubsection{Mitt Romney}
\paragraph{Policies-Interests Score: 4/5}
\begin{itemize}
\item Lower tax rates for individuals and corporations: Proposed across-the-board tax rate reductions.
\item Achieve energy independence: Emphasized increasing domestic oil and gas production.
\item Open up more trade, especially in Latin America: Advocated for more free trade agreements.
\item Improve education and job training: Proposed grading schools and allowing school choice.
\item Balance the budget: Stressed reducing government spending.
\item Support small businesses: Positioned as champion of small businesses.
\item Repeal and replace Affordable Care Act: Pledged to repeal "Obamacare".
\end{itemize}
\paragraph{Persona-Identity Score: 4/5}
\begin{itemize}
\item Successful businessman: Referenced private sector experience.
\item Problem-solver: Presented as practical, results-oriented leader.
\item Washington outsider: Emphasized background outside of Washington.
\item Defender of free enterprise: Advocated for free market solutions.
\end{itemize}
\paragraph{Perspective-Ideologies Score: 4/5}
\begin{itemize}
\item Conservative: Policy proposals aligned with traditional conservative values.
\item Advocate for limited government: Argued for reducing size and scope of federal government.
\item Supporter of free market solutions: Expressed faith in private sector's problem-solving ability.
\item Proponent of states' rights: Emphasized importance of state-level decision-making.
\end{itemize}
\textbf{Romney's LLM-POTUS Score:} $\frac{4 + 4 + 4}{3} = 4.00$
\begin{table}[h]
\centering
\begin{tabular}{|l|c|c|}
\hline
\textbf{Aspect} & \textbf{Barack Obama} & \textbf{Mitt Romeney} \\
\hline
Policies - Interests & 4/5 & 4/5 \\
Persona - Identity & 4/5 & 4/5 \\
Perspective - Ideologies & 5/5 & 4/5 \\
\hline
\textbf{Average Score} & 4.33 & 4.00 \\
\hline
\end{tabular}
\caption{2012 LLM-POTUS Claude3 Scores Comparison}
\end{table}

\section{Discussions}
\subsection{Impact}
This study represents a pioneering approach in utilizing large language models as independent judges to assess presidential debates. While not entirely objective, LLMs provide a novel perspective for evaluating debate performances, which potentially complements traditional human analysis with insights derived from vast language understanding capabilities. This method's scalability also allows for rapid analysis of multiple debates across various election cycles. Indeed, large-scale comparative studies are time-consuming for human analysts. 

It is important to note that existing methods of analyzing U.S. presidential debates come with their own set of biases. Election polls can be influenced by sampling errors, response biases, and the phrasing of questions~\cite{crespi1988pre,gelman1993american}. Human analysts, despite their expertise, may have personal or ideological biases that color their interpretations~\cite{d2000media,brubaker2009effect,gross2017presidential}. Similarly, media coverage of debates often reflects the political leanings of news organizations or the preferences of their audience~\cite{kuypers2002press,d2000media}. In this context, our LLM-POTUS Score offers an alternative approach that, while not free from biases inherited from model training, provides an alternative and flexible form of analysis that is less susceptible to direct human subjectivity or blatant media influence.

The LLM-POTUS Score introduces a multidimensional assessment that goes beyond policy positions and incorporates factors such as candidate persona and ideological alignment in a structured manner. This approach may reveal patterns or trends in debate performances that human analysts might overlook and offers fresh perspectives on political communication strategies. By providing a standardized, AI-driven analysis, this method opens new avenues for understanding the nuances of political discourse and candidate performance in high-stakes debates.

Furthermore, this framework has significant civic implications, as it provides individual citizens with an independent tool to evaluate debate performances, which can enhance democratic engagement by reducing reliance on potentially biased media interpretations and external analysts. This democratization of political analysis has the potential to foster a more informed and critically engaged electorate.

\subsection{Limitations}
Despite its potential, our approach faces several limitations that warrant careful consideration. The AI models used may inherit biases present in their training data, potentially skewing their analysis of certain candidates or issues~\cite{yeh2023evaluating}. There is also a risk of hallucination~\cite{mcdonald2024reducing}, where LLMs might generate plausible-sounding but factually incorrect information, which could lead to erroneous analyses if not carefully monitored. 

Our current method also relies solely on text transcripts, which do not include crucial non-verbal cues such as gestures, facial expressions, tone of voice, and overall appearance. These factors play a significant role in debate performances. The temporal context is another limitation, as LLMs' knowledge cutoff dates may restrict their understanding of the most recent political developments or changes in public opinion. 

Additionally, the variation in results between different LLMs raises questions about consistency and which model's analysis should be considered authoritative. These limitations highlight the need for cautious interpretation of results and ongoing refinement of the methodology to enhance its reliability and comprehensiveness in political debate analysis.

\subsection{Future Perspectives}
\subsubsection{Multimodal and Tailored Models}
A direction for future research lies in the development of specialized multimodal models that integrate video and audio data alongside text transcripts. Such models could potentially allow for a more comprehensive analysis of presidential debates, encompassing non-verbal communication, appearance, and emotional cues. This approach may provide a richer understanding of debate dynamics beyond what is captured in text alone.

Furthermore, the fine-tuning of language models or multimodal models specifically for political discourse analysis represents another avenue for exploration. Such tailored models could potentially improve accuracy and reduce biases inherent in general-purpose models. This specialized approach may yield more nuanced insights into the complexities of political communication during debates.

\subsubsection{Real-time Audience Reaction Integration}
The current method lacks the ability to account for immediate audience responses or the dynamic interplay between candidates and viewers during live debates. Future iterations should incorporate data on real-time audience reactions (e.g., social media sentiment, focus group responses) to provide a more holistic view of debate impact and capture the immediate public response to candidates' performances.

\subsubsection{Comparative Studies}
Future research could benefit from extensive comparisons between LLM-based analyses and human expert evaluations. Such studies may serve to validate the method and identify areas of divergence or complementarity. This comparative approach could provide valuable insights into the strengths and limitations of AI-driven political analysis vis-à-vis traditional human expertise.

\subsubsection{Historical Analysis}
An interesting avenue for exploration is the application of the LLM-POTUS Score methodology to historical debates. This approach may uncover long-term trends in political communication strategies, offering a unique perspective on the evolution of political discourse over time. However, careful consideration must be given to potential biases in historical data and the changing landscape of political representation.

\subsubsection{Cross-cultural Application}
The potential for cross-cultural application of this methodology presents another promising direction for future research. Adapting the LLM-POTUS Score for use in analyzing political debates in other countries could yield valuable comparative insights. Such adaptations would necessitate careful consideration of cultural and political differences, potentially enriching our understanding of global political communication patterns.

\subsubsection{Ethical Framework}
As AI continues to play an increasing role in political analysis~\cite{wu2023large,linegar2023large,rodman2024political}, the development of a robust ethical framework becomes paramount. Future work in this area could focus on establishing clear guidelines for the ethical use of AI in political analysis. Such a framework would aim to ensure transparency and address concerns about AI's influence on democratic processes, potentially including considerations for public accessibility of the methodology and criteria.

\subsubsection{Weighted Average Scoring}
One direction for future research is the implementation of a weighted average for the LLM-POTUS Score, as opposed to the current simple average of the three metrics. We can consider developing a methodology to learn these weights through analysis of voter priorities regarding policies, persona, and perspectives. Such an approach could potentially provide a more nuanced and accurate representation of debate performance impact on different voter segments.

\subsubsection{Election Outcome Prediction}
The potential of the LLM-POTUS Score as a factor in predicting election outcomes represents another avenue for exploration. Currently, election forecasting relies heavily on polling data~\cite{kenett2018election} and socioeconomic indicators~\cite{leighley2013votes}. Therefore, integrating this score with other traditional predictive models could offer a new dimension in election forecasting, combining AI-driven debate analysis with established methodologies. This integration may provide valuable insights into the relationship between debate performance and electoral success.

\subsubsection{Fact-Checking Enhanced Model}
Currently, the LLM-POTUS methodology relies solely on the inherent knowledge of large language models. Therefore, a promising direction for improvement is the development of a dedicated model that integrates fact-checking capabilities or utilizes retrieval-augmented generation (RAG) techniques~\cite{wu2024retrieval}. Such enhancements could significantly reduce the risk of hallucinations and increase the accuracy of the analysis. For instance, real-time fact-checking tools like ClaimBuster\footnote{\url{https://idir.uta.edu/claimbuster/}} could be incorporated to verify claims, potentially adding an "Integrity" dimension to the 3P alignment and producing more reliable candidate assessments.

\section{Conclusion}
This study introduces the LLM-POTUS Score, a novel framework for analyzing presidential debates using large language models. By evaluating candidates' performances across the dimensions of Policies, Persona, and Perspective, and their alignment with audience Interests, Identities, and Ideologies, our method offers a nuanced, multi-dimensional assessment of debate effectiveness. The application of this framework to U.S. presidential debates from 2000 to 2024 demonstrates its potential to provide consistent, scalable analysis across multiple election cycles. While there are limitations such as potential biases in training data and the exclusion of non-verbal cues, this approach represents a step towards independent and objective political discourse analysis that is accessible to every citizen. As AI continues to evolve, the LLM-POTUS Score methodology opens new avenues for understanding political communication, potentially complementing traditional analysis methods and contributing to a more comprehensive evaluation of presidential debates. Future research directions, including multimodal analysis, real-time audience integration, and cross-cultural applications, promise to further refine and expand the utility of this innovative approach in political science.

\bibliography{LLM_refs}
\bibliographystyle{unsrt}

\end{document}